\documentclass[pdftex]{article}

\usepackage[final, nonatbib]{neurips_2023}
\usepackage[square,sort,comma,numbers]{natbib}
\usepackage{lipsum}
\usepackage{algorithm}
\usepackage{multirow}
\usepackage{algpseudocode}
\usepackage{amsmath}
\usepackage{wrapfig}
\usepackage{amssymb}
\usepackage{natbib}
\usepackage{soul}
\usepackage{algorithm}
\usepackage{adjustbox}

\usepackage[utf8]{inputenc} 
\usepackage[T1]{fontenc}    
\usepackage{hyperref}       
\usepackage{url}            
\usepackage{booktabs}       
\usepackage{amsfonts}       
\usepackage{nicefrac}       
\usepackage{microtype}      
\usepackage{xcolor}         
\usepackage[pdftex]{graphicx} 
\usepackage{microtype}
\usepackage{subfigure}
\usepackage{booktabs} 
\usepackage{color,colortbl}
\usepackage[linewidth=1pt]{mdframed}
\usepackage{framed}
\usepackage{adjustbox}

\usepackage{amsmath}
\usepackage{amssymb}
\usepackage{mathtools}
\usepackage{amsthm}

\usepackage[capitalize,noabbrev]{cleveref}

\theoremstyle{plain}

\theoremstyle{definition}

\theoremstyle{remark}

\usepackage[textsize=tiny]{todonotes}

\title{Fast Fourier Transform-Based Spectral and Temporal Gradient Filtering for Differential Privacy}

\author{
  Hyeju Shin\textsuperscript{1,5,\dag}, 
  Vincent-Daniel Yun\textsuperscript{2,5,\dag}\thanks{Currently Under Review. Corresponding author: \texttt{juyoung.yun@usc.edu}},
  Kyudan Jung\textsuperscript{3,5,\dag}, 
  Seongwon Yun\textsuperscript{4,5,\dag} \\ \\
  \textsuperscript{1}Electronics and Telecommunications Research Institute (ETRI), Republic of Korea \\
  \textsuperscript{2}University of Southern California, Viterbi, United States \\
  \textsuperscript{3}KAIST, AI, Republic of Korea \\
  \textsuperscript{4}Hanwha Life Insurance, Republic of Korea \\
  \textsuperscript{5}OpenNN Lab, MODULABS, Republic of Korea \\ \\
  \textsuperscript{\dag}Equal contribution
}

\begin{document}

\maketitle

\let\thefootnote\relax\footnotetext{}

\begin{abstract}
Differential Privacy (DP) has emerged as a key framework for protecting sensitive data in machine learning, but standard DP-SGD often suffers from significant accuracy loss due to injected noise. To address this limitation, we introduce the FFT-Enhanced Kalman Filter (FFTKF), a differentially private optimization method that improves gradient quality while preserving $(\varepsilon, \delta)$-DP guarantees. FFTKF applies frequency-domain filtering to shift privacy noise into less informative high-frequency components, preserving the low-frequency gradient signals that carry most learning information. A scalar-gain Kalman filter with a finite-difference Hessian approximation further refines the denoised gradients. The method has per-iteration complexity $\mathcal{O}(d \log d)$ and achieves higher test accuracy than DP-SGD and DiSK on MNIST, CIFAR-10, CIFAR-100, and Tiny-ImageNet with CNNs, Wide ResNets, and Vision Transformers. Theoretical analysis shows that FFTKF ensures equivalent privacy while delivering a stronger privacy--utility trade-off through reduced variance and controlled bias.

\end{abstract}


\section{Introduction}
Differential Privacy (DP) has become a foundational framework for safeguarding individual-level information in machine learning.

\begin{wrapfigure}{l}{0.5\textwidth}
    \centering
    \vspace{-15pt}
    \includegraphics[width=\linewidth]{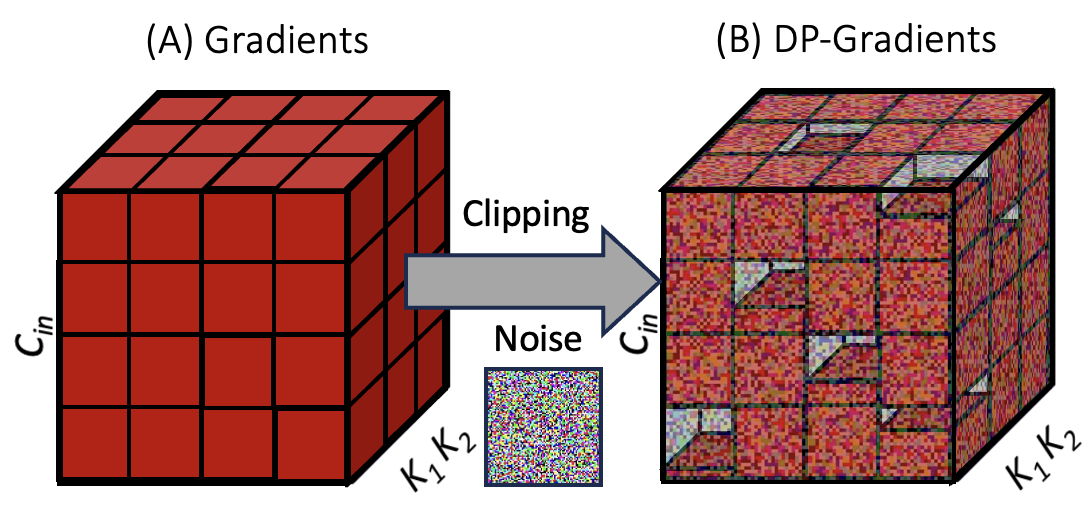}
    \caption{Illustration of the Differentially Private Stochastic Gradient Descent (DP-SGD) process. (A) Original gradients (B) DP-Gradients}
    \label{fig:dpsgd}
    \vspace{-15pt}
\end{wrapfigure} 
This provides rigorous guarantees against information leakage from model outputs~\citep{abadi2016deep, wang2024adaptivedifferentialprivacymethod, selvi2024differentialprivacydistributionallyrobust, lomurno2022utilityprotectionoptimizationdifferential}.

Standard DP mechanisms, such as the Laplace and Gaussian mechanisms, achieve privacy by injecting calibrated noise into data or gradients, as shown in Figure~\ref{fig:dpsgd}. However, this noise often causes significant degradation in model utility, especially in high-dimensional or deep models.

A central challenge in DP learning is improving the performance of DP-SGD~\citep{abadi2016deep}. The high variance of DP noise under tight privacy budgets leads to poor signal-to-noise ratios, slowing convergence and reducing accuracy~\citep{pichapati2019adaclip, shulgin2024convergencedpsgdadaptiveclipping}. Thus, denoising while preserving $(\epsilon,\delta)$-DP remains an open problem.

To address this, recent works integrate signal processing and state estimation into DP optimization. The DiSK framework~\citep{zhang2025disk} applies Kalman filtering to iteratively estimate cleaner gradients, leveraging temporal correlations~\citep{ma2022kalman, ny2012differentiallyprivatekalmanfiltering}. In parallel, frequency-domain methods use low-pass filtering to separate useful gradient signals from high-frequency noise~\citep{Eldar_2010, liu2020random, ordonez2022adaptive, fourier2025optimization, andrew2022differentiallyprivatelearningadaptive, shulgin2024convergencedpsgdadaptiveclipping}. These advances suggest that combining temporal and spectral denoising can substantially improve the utility of privatized gradients.

Building on this, we propose the \textit{FFT-Enhanced Kalman Filter} (FFTKF), which reshapes DP noise into high-frequency components via Fast Fourier Transform (FFT) and then applies a scalar-gain Kalman filter to recover stable low-frequency gradients. This approach preserves $(\epsilon,\delta)$-DP while improving convergence and test accuracy.

\begin{framed}
\noindent
\textbf{Our contributions are summarized as follows:}
\begin{itemize}
    \item A frequency-domain noise shaping strategy that retains DP guarantees.
    \item A lightweight Kalman filter update with per-step complexity $O(d \log d)$.
    \item Empirical validation on MNIST, CIFAR-10, CIFAR-100, and Tiny-ImageNet across CNNs, Wide ResNets, and Vision Transformers, showing consistent gains over DP-SGD and DiSK.
\end{itemize}
\end{framed}

\section{Related Works}
Stochastic Gradient Descent (SGD) and its variants such as Adam are the backbone of modern optimization~\citep{robbins1951stochastic, kingma2014adam}. SGD provides efficiency by using mini-batch gradients but suffers from high variance, while Adam improves stability through momentum and adaptive scaling. Despite their success, these methods offer no inherent privacy, as gradients may expose sensitive data. This motivated the development of privacy-preserving optimizers such as DP-SGD~\citep{abadi2016deep}. 

Differential Privacy (DP) ensures rigorous protection by injecting calibrated noise into data or gradients~\citep{abadi2016deep, wang2024adaptivedifferentialprivacymethod, selvi2024differentialprivacydistributionallyrobust, lomurno2022utilityprotectionoptimizationdifferential}. DP-SGD achieves $(\epsilon,\delta)$-DP through Gaussian perturbation but often reduces utility under tight budgets~\citep{abadi2016deep}. To mitigate this, adaptive noise adjustment has been explored~\citep{fan2013adaptive}, and the DiSK framework introduced Kalman filtering for denoising~\citep{zhang2025disk}. Adaptive clipping further improves learning by tuning gradient norms dynamically~\citep{thakkar2019differentially}. Together, these works highlight the challenge of balancing privacy and accuracy. 

Kalman filters estimate hidden states in noisy systems by leveraging temporal correlations~\citep{ma2022kalman, ny2012differentiallyprivatekalmanfiltering}. In DP optimization, DiSK applies a simplified Kalman filter to stabilize noisy gradient updates, improving convergence with low computational cost~\citep{zhang2025disk}. This temporal smoothing has proven useful in large-scale models and has also been extended to federated learning, where client updates require both privacy and accuracy. These studies show that Kalman-based methods are flexible tools for gradient denoising under DP constraints. Low-pass filters suppress high-frequency noise while preserving dominant low-frequency signals~\citep{fourier2025optimization, bracewell1999fourier}. Fourier-based approaches are especially attractive due to their computational efficiency. In DP, adaptive low-pass filtering has been proposed to maximize utility while meeting privacy budgets, effectively recovering gradient information~\citep{andrew2022differentiallyprivatelearningadaptive, shulgin2024convergencedpsgdadaptiveclipping, chen2023differentiallyprivate}. These techniques are particularly relevant in deep learning, where most useful gradient information lies in low-frequency components. \\

\section{Methodology}
\noindent\textbf{Preliminaries.}
We work in $\mathbb{R}^d$ with $\ell_2$–norm $\|\cdot\|_2$.
Let $I_d$ be the identity and $\mathrm{diag}(\varphi_0,\ldots,\varphi_{d-1})$ a diagonal matrix. For $f:\mathbb{R}^d\!\to\!\mathbb{R}$, denote gradient and Hessian by $\nabla f$ and $\nabla^2 f$.
We use the Hadamard product $\odot$, the floor $\lfloor\cdot\rfloor$, and the Gaussian law $\mathcal{N}(0,\sigma^2 I_d)$. We minimize the population loss $F(x)=\mathbb{E}_{\xi\sim\mathcal{D}}[f(x;\xi)]$ via iterates
$x_{t+1}=x_t-\eta\,\tilde g_t$ with step size $\eta>0$.
Given a mini-batch $\mathcal{B}_t$ of size $B$, the stochastic gradient is
$g_t=\tfrac{1}{B}\sum_{\xi\in\mathcal{B}_t}\nabla f(x_t;\xi)$, and we write the parameter difference as $d_t=x_{t+1}-x_t$. A mechanism $\mathcal{M}$ is $(\varepsilon,\delta)$-DP if for any neighboring datasets $D,D'$ and event $S$,
$\Pr[\mathcal{M}(D)\in S]\le e^\varepsilon \Pr[\mathcal{M}(D')\in S]+\delta$.
In DP-SGD, per-sample gradients are clipped $\mathrm{clip}(v,C)=v\cdot \min\!\bigl(1,\,C/\|v\|_2\bigr)$ and Gaussian noise $w_t\!\sim\!\mathcal{N}(0,\sigma_w^2 I_d)$ is added to enforce privacy.

\begin{figure*}
    \centering
    \includegraphics[width=0.995\linewidth]{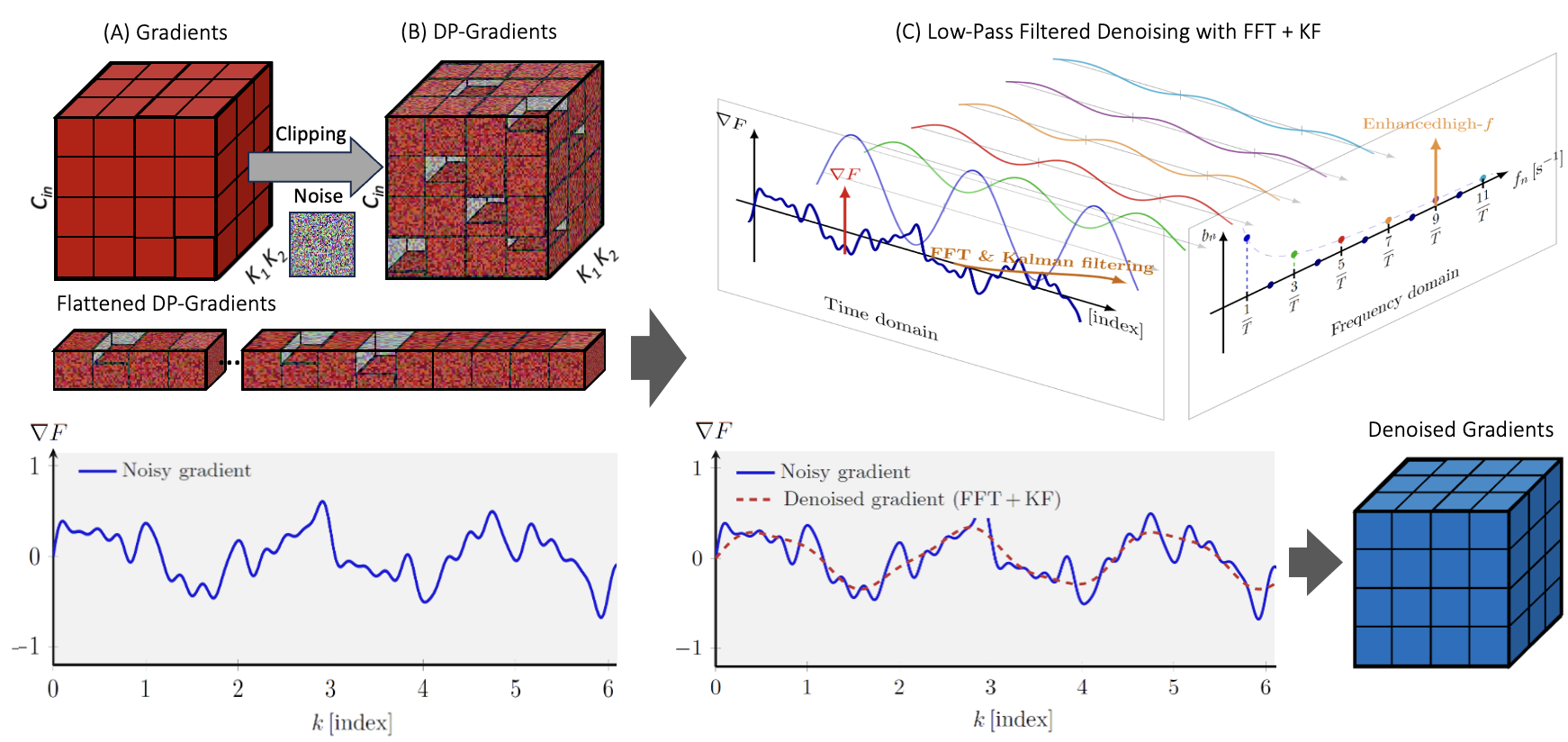}
    \vspace{-12pt}
    \caption{%
    Visualization of the proposed frequency-domain gradient denoising process. 
    (A) Original gradients before privatization. 
    (B) Differentially private gradients obtained by clipping and adding Gaussian noise. 
    (C) Our FFT+Kalman filtering method denoises the privatized gradients in the frequency domain, reducing high-frequency perturbations while preserving the underlying signal structure.%
    }
    \label{fig:enter-label}
\end{figure*}

\subsection{Fast Fourier Transform}
\label{ssec:fft_recap}
This section briefly reviews the discrete Fast Fourier transform (FFT) and the
algorithmic considerations that motivate its use for gradient denoising. Recall that the \emph{discrete Fourier transform} (DFT) of a real-valued vector \( z = (z_0, \dots, z_{d-1})^\top \in \mathbb{R}^d \) is the complex vector \( \hat{z} = \mathcal{F}(z) \in \mathbb{C}^d \), with components $\hat{z}_k = \sum_{n=0}^{d-1} z_n \, e^{-2\pi i kn/d}$ when $k = 0, \dots, d-1$ and its inverse is:
\begin{align}
z_n &= \frac{1}{d} \sum_{k=0}^{d-1} \hat{z}_k \, e^{2\pi i kn/d}, \quad n = 0, \dots, d-1.
\end{align}

With this normalization, the Fourier transform $\mathcal{F}$ is unitary is $\mathcal{F}^{-1}(\mathcal{F}(z)) = z$ and Parseval’s identity holds:
\begin{align}
\|z\|_2^2
&= z^\ast z \\
&= z^\ast \, \mathcal{F}^{-1} \mathcal{F} \, z \\
&= (\mathcal{F} z)^\ast \, (\mathcal{F} z) \cdot \frac{1}{d} \\
&= \frac{1}{d} \|\hat{z}\|_2^2. \label{eq:parseval_expansion}
\end{align}
where $\hat{z} = \mathcal{F}(z)$.

Consequently, injecting Gaussian noise in the Fourier domain preserves the $\ell_2$-sensitivity required for $(\varepsilon,\delta)$-DP, since the transform is unitary and does not amplify vector norms.
\medskip

\noindent\textbf{Low/high–frequency split.}
Fix a pivot index \(k_0=\lfloor \lambda d\rfloor\) for some \(\lambda\in(0,1)\). Frequencies \(k<k_0\) are called \emph{low-frequency components}, and \(k\ge k_0\) \emph{high-frequency components}. This separation reflects the empirical observation that most signal information, especially in gradient vectors of smooth loss landscapes, is concentrated in the lower spectral range, while the high-frequency components often contain stochastic noise.

\medskip




\noindent\textbf{Spectral filtering.}
A diagonal mask \( \Phi = \mathrm{diag}(\varphi_0, \dots, \varphi_{d-1}) \) defines a linear filter $\mathcal{G}_\Phi(z) = \mathcal{F}^{-1} \left( \Phi\, \hat{z} \right)
= \frac{1}{d} \sum_{k=0}^{d-1} \varphi_k \hat{z}_k \, e^{2\pi i kn/d}$ where \( \hat{z} = \mathcal{F}(z) \). Equivalently, in matrix and convolution form,
\begin{align}
\mathcal{G}_\Phi &= \mathcal{F}^{-1}\Phi \mathcal{F}, \\
(\mathcal{G}_\Phi z)_n &= \sum_{m=0}^{d-1} h_{(n-m)\bmod d}\,z_m, \quad 
h = \mathcal{F}^{-1}\varphi.
\end{align}
By Parseval’s identity,
\begin{align}
\|\mathcal{G}_\Phi z\|_2^2
= \tfrac{1}{d}\sum_{k=0}^{d-1} |\varphi_k|^2\,|\hat z_k|^2
\le (\max_k |\varphi_k|^2)\,\|z\|_2^2.
\end{align}
By the convolution theorem, this operation in the frequency domain is equivalent to convolution in the time domain and can be evaluated in \(O(d \log d)\) time via the FFT algorithm, which significantly improves efficiency compared to the naive \(O(d^2)\) convolution \citep{hazan2018spectralfilteringgenerallinear, briggs1995dft, tolimieri1997algorithms}.
\medskip

\noindent\textbf{High-frequency shaping mask.}
To enhance denoising while maintaining DP, we use a smooth mask function
\begin{align}
  \varphi_k =
  \begin{cases}
    1, & k < k_0, \\[3pt]
    1 - \rho, & k \geq k_0 .
  \end{cases}
\end{align}

where $\rho \in (0, 1)$ controls the magnitude of suppression. This *step-wise attenuation* suppresses higher-frequency components beyond a cutoff index $k_0$, thereby reducing the influence of DP noise concentrated in those frequencies. Unlike sharp cutoffs, this mask gently dampens high-frequency content while preserving the low-frequency structure of gradients, offering a balance between denoising and signal fidelity \citep{zhang2024doppler,wang2024adaptivedifferentialprivacymethod}.

\medskip

Let \(\Phi_\rho = \mathrm{diag}(\varphi_0, \dots, \varphi_{d-1})\), then the filtered version of a privatized gradient \(g = \nabla f + w\) is $\hat{g} = \mathcal{G}_{\Phi_\rho}(g) = \mathcal{F}^{-1}(\Phi_\rho \mathcal{F}(g)).$
When \(w \sim \mathcal{N}(0, \sigma^2 I)\), the transformed noise \(\hat{w} := \mathcal{F}^{-1}(\Phi_\rho \mathcal{F}(w))\) is still zero-mean but now has reduced energy in the low-frequency components:
\begin{align}
  \mathbb{E}[\|\hat{w}_{<k_0}\|_2^2] \ll \mathbb{E}[\|\hat{w}\|_2^2],
\end{align}
facilitating more accurate recovery of the gradient signal after filtering.

\medskip

This FFT recap underpins our \textit{FFT-Enhanced Kalman Filter} in Sec.~\ref{ssec:fftkf}, where we combine spectral noise shaping with a scalar-gain Kalman predictor to denoise privatized gradients efficiently, achieving both computational and privacy-preserving benefits.

\subsection{Gradient Dynamics with High-Frequency Differential Privacy}
\label{ssec:dynamics}
To explain our proposed idea of using the FFT-Enhanced Kalman Filter for denoising gradients, we first establish a dynamic system for the gradients. This system consists of a \emph{system update} equation and an \emph{observation} equation. The system update of the gradient dynamics is derived via Taylor expansion of $\nabla F$ around $x_{t-1}$, allowing for a second-order approximation of the gradient evolution at step $t$:
\begin{align}
\nabla F(x_t)
&= \nabla F(x_{t-1}+d_{t-1}) \\[3pt]
&= \nabla F(x_{t-1})
   + \nabla^2F(x_{t-1})\,d_{t-1} \\[3pt]
&\quad + \tfrac{1}{2}\int_0^1
   \nabla^3F\bigl((1-z)x_{t-1}+z x_t\bigr)[d_{t-1}]^{\otimes 2}\,dz,
\label{eq:sysupdate}
\end{align}
where $\boldsymbol{H}_t := \nabla^2F(x_{t-1}) \in \mathbb{R}^{d \times d}$ is approximated using privatized finite differences, and $d_{t-1}=x_t-x_{t-1}$.
\medskip
\noindent The observed gradient $g_t$ is a noisy, privatized estimate of the true gradient:
\begin{align}
g_t
&= \frac{1}{B}\sum_{\xi \in \mathcal{B}_t}\text{clip}(\nabla f(x_t, \xi), C) + w_t \\[3pt]
&= C_t \nabla F(x_t) + w'_t,
\end{align}
where $w'_t$ contains both DP noise and subsampling noise, and $C_t$ is the effective observation operator with $\|C_t\|_2 \leq 1$.
\medskip
\noindent Combining the update and observation equations:
\begin{align}
\nabla F(x_t)
&= \nabla F(x_{t-1}) + \boldsymbol{H}_t (x_t - x_{t-1}) + v_t, \tag{System update} \\[3pt]
g_t
&= C_t \nabla F(x_t) + w'_t. \tag{Observation}
\end{align}
\medskip
\noindent To enforce differential privacy while retaining useful structure, we first apply isotropic Gaussian noise $w_t \sim \mathcal{N}(0, \sigma_w^2 I_d)$ to the clipped gradient, followed by a deterministic frequency-domain transformation to shape the noise:
\begin{align}
g_t
&= C_t \nabla F(x_t) + w_t', \\[3pt]
w_t'
&= \mathcal{F}^{-1} \bigl( \Phi_\rho \odot \mathcal{F}(w_t) \bigr),
\label{eq:obs}
\end{align}
where $\Phi_\rho \in \mathbb{R}^d$ satisfies
\begin{align}
(\Phi_\rho)_k
&=
\begin{cases}
  1, & 0 \le k < k_0, \\[3pt]
  1 - \rho e^{-\alpha(k - k_0)}, & k_0 \le k < d,
\end{cases}
\end{align}
with $k_0 = \lfloor \lambda d \rfloor$, $\rho \in (0,1)$, and $\alpha > 0$. This ensures that the privacy-preserving noise $w_t'$ is spectrally shaped to occupy primarily high-frequency components, which contribute less to gradient descent, while preserving the $(\varepsilon, \delta)$-DP guarantee through post-processing. This approach facilitates improved recoverability of the informative low-frequency gradient content.

\subsection{Frequency-Domain Denoising}
\label{ssec:fftfilter}
To recover the low-frequency content of the privatized gradient, we apply the inverse of the noise shaping operation:
\begin{align}
  \mathcal{G}_{\rho}(z)
  &:=
  \mathcal{F}^{-1}\!\bigl(
     \Phi_\rho \odot \mathcal{F}(z)
  \bigr),\qquad
  z \in \mathbb{R}^d.
\end{align}
This filtering step yields the estimate \(\widehat{g}_t = \mathcal{G}_{\rho}(g_t)\). Since \(\mathcal{G}_{\rho}\) is a linear operator with spectral mask \(\Phi_\rho\), this operation has complexity \(O(d \log d)\) and does not distort the signal beyond a known attenuation factor. The covariance of \(\widehat{g}_t\) is a spectrally reweighted version of \(\mathrm{Cov}(g_t)\), which we exploit in the Kalman update below \citep{Kasanick__2015}.

\subsection{FFT-Enhanced Kalman Filter}
\label{ssec:fftkf}
We adopt the scalar-gain Kalman filtering approximation introduced in \cite{zhang2025disk}, which simplifies the covariance matrices to scalar multiples of the identity. Specifically, we let \(P_t = p_t I_d\), \(K_t = \kappa I_d\), and estimate the Hessian action using a privatized finite-difference formula with hyperparameter \(\gamma > 0\).
\medskip

\noindent\textbf{Prediction Step.}
Given \(\tilde{g}_{t-1}\), we predict the next gradient by using a first-order approximation based on privatized finite differences:
\begin{align}
  \tilde{g}_{t|t-1} = 
     &\tilde{g}_{t-1} \\
     &+\frac{1}{B}\sum_{\xi \in \mathcal{B}_t}
       \frac{\text{clip}(\nabla f(x_t + \gamma d_{t-1}; \xi), C)}{\gamma} \\
       &- \sum_{\xi \in \mathcal{B}_t}\frac{\text{clip}(\nabla f(x_t; \xi), C)}{\gamma} + w_t^\text{fd},
     \label{eq:prediction}
\end{align}
where \(d_{t-1} := x_t - x_{t-1} = -\eta \tilde{g}_{t-1}\), \(w_t^\text{fd} \sim \mathcal{N}(0, \sigma_\text{fd}^2 I_d)\) is additional noise for privacy, and clipping is applied to bound sensitivity. This approximates the action of the local Hessian without explicitly computing second-order derivatives.
\medskip

\noindent\textbf{Correction Step.}
The predicted gradient is then corrected using the filtered observation \(\widehat{g}_t\):
\begin{equation}
  \tilde{g}_t =
     (1 - \kappa) \tilde{g}_{t-1}
     + \kappa \widehat{g}_t,
     \tag{C}
     \label{eq:correction}
\end{equation}
where \(\kappa \in (0,1)\) is the Kalman gain that balances the reliance on the prediction versus the new (denoised) observation. This form ensures that the update direction incorporates temporal consistency across iterations while attenuating the influence of high-frequency noise.
\medskip
Together, Eqs.~\eqref{eq:prediction} and \eqref{eq:correction} constitute a computationally lightweight Kalman filtering mechanism enhanced by frequency-domain denoising. The per-step complexity is \(O(d \log d)\) for FFT operations plus \(O(d)\) for two gradient evaluations and finite-difference computation, achieving overall efficiency while enhancing DP optimization without sacrificing privacy guarantees.
\label{ssec:algorithm}
\begin{algorithm}[htb!]
  \caption{FFT-Enhanced Kalman Filter Optimizer (FFTKF)}
  \label{alg:fftkf}
  \begin{algorithmic}[1]
    \Require initial point $x_0$, base optimiser $\mathsf{Opt}$,
            learning rate $\eta$, gain $\kappa$, FD parameter $\gamma$,
            high–frequency ratio $\rho$, clipping bound $C$, noise scales $\sigma_w$, $\sigma_\text{fd}$.
    \State $\tilde{g}_{-1} \gets 0$, $d_{-1} \gets 0$
    \For{$t = 0, 1, \dots, T-1$}
      \State Sample mini-batch $\mathcal{B}_t$
      \State Compute privatized gradient with isotropic noise
        \begin{align}
           g_t
           \leftarrow
           \frac{1}{B} \sum_{\xi \in \mathcal{B}_t} \operatorname{clip}\bigl(\nabla f(x_t; \xi), C\bigr)
           + w_t, \quad w_t \sim \mathcal{N}(0, \sigma_w^2 I_d)
        \end{align}
      \State $\widehat{g}_t \leftarrow \mathcal{G}_{\rho}(g_t)$ \Comment{FFT denoising}
      \State $\tilde{g}_{t|t-1} \leftarrow$ Eq.~\eqref{eq:prediction} \Comment{Privileged finite-difference prediction}
      \State $\tilde{g}_t \leftarrow$ Eq.~\eqref{eq:correction}
      \State $x_{t+1} \leftarrow \mathsf{Opt}\bigl(x_t, \eta, \tilde{g}_t\bigr)$
      \State $d_t \gets x_{t+1} - x_t$
    \EndFor
  \end{algorithmic}
\end{algorithm}

\subsection{Additional Discussion}
The high-frequency shaping in Eq.~\eqref{eq:obs} intentionally pushes
privacy noise into spectral regions that matter least for optimization.
Because the Kalman filter relies on low-frequency temporal correlations captured by
Eqs.~\eqref{eq:prediction}–\eqref{eq:correction}, the FFT step removes most of
the injected disturbance before the gain $\kappa$ is applied, resulting in a
provably lower steady-state covariance.

Let \(\Sigma_w = \sigma_w^2 I_d\) be the covariance of the original DP noise \(w_t\), then the shaped noise \(\tilde{w}_t = \mathcal{F}^{-1}(\Phi_\rho \odot \mathcal{F}(w_t))\) has covariance
\begin{align}
  \Sigma_{\tilde{w}} = \mathcal{F}^{-1} \cdot \Phi_\rho^2 \cdot \mathcal{F} \cdot \Sigma_w \cdot \mathcal{F}^{-1} \cdot \Phi_\rho^2 \cdot \mathcal{F},
\end{align}
whose low-frequency principal components are suppressed relative to \(\Sigma_w\). Hence, the Kalman filter receives observations with diminished low-frequency noise variance, resulting in lower mean-square estimation error.

Crucially, FFTKF inherits the \emph{$O(d)$ memory} and \emph{$O(d)$ algebraic}
complexities of the simplified DiSK variant while adding only two
in-place FFTs per iteration. \\

\noindent \textbf{Scalar–gain Kalman simplification.} Our FFT-Enhanced Kalman Filter (\textit{FFTKF}) inherits the scalar–gain reduction of \textsc{DiSK}~\cite{zhang2025disk}, wherein both the state covariance \(P_t\) and the Kalman gain \(K_t\) are isotropic:
\begin{align}
  P_t = p_t I_d, \quad K_t = \kappa I_d.
\end{align}
This diagonal simplification ensures that all matrix-vector operations reduce to scalar multiples of vector additions, preserving an \(\mathcal{O}(d)\) runtime and storage profile. The Hessian-vector product \(\boldsymbol{H}_t d_{t-1}\) is approximated with a single finite-difference query:
\begin{align}
  \boldsymbol{H}_t d_{t-1} \approx \frac{\nabla F(x_t + \gamma d_{t-1}) - \nabla F(x_t)}{\gamma},
\end{align}
eliminating the need for Hessian storage or inversion. \\

\noindent \textbf{FFT-based noise shaping.} While \textsc{DiSK} performs time-domain exponential smoothing, FFTKF additionally \emph{reshapes} the injected DP noise to concentrate its energy in the high-frequency spectrum:
\begin{align}
  \begin{aligned}
    \tilde w_t &= \mathcal{F}^{-1}\!\bigl(\Phi_\rho \odot \mathcal{F}(w_t)\bigr),
  \end{aligned}
\text{ }
  \begin{aligned}
    (\Phi_\rho)_k &=
    \begin{cases}
      1, & k < k_0, \\[4pt]
      1 - \rho e^{-\alpha(k-k_0)}, & k \ge k_0 .
    \end{cases}
  \end{aligned}
\end{align}

with pivot index \(k_0 = \lfloor \lambda d \rfloor\). The mask \(\Phi_\rho\) acts as a soft high-pass filter for the noise, minimizing the effect of noise on low-frequency directions where the Kalman filter’s predictive prior is most accurate. This filtering can be viewed as a dual to the temporal smoothing in DiSK, but operating in the spectral domain. \\

\noindent \textbf{Computational footprint.} Compared with DPSGD, FFTKF requires one additional forward pass per iteration to compute the finite-difference directional gradient, a forward transform \(\mathcal{F}\), and its inverse \(\mathcal{F}^{-1}\). Both operations scale as \(O(d \log d)\), while the state vector \(\tilde g_t\) and difference direction \(d_t\) are stored as \(O(d)\) vectors. Thus, FFTKF matches the memory profile of DiSK~\cite{zhang2025disk} but enables more precise noise attenuation with marginal overhead. \\

\noindent \textbf{Privacy guarantee.}
Since the FFT operation is orthonormal, it preserves the \(\ell_2\) norm:
\begin{align}
  \|\tilde{w}_t\|_2 = \|\Phi_\rho \odot \mathcal{F}(w_t)\|_2 \le \|w_t\|_2.
\end{align}
Thus, FFT-based reshaping does not increase the sensitivity of the privatized quantity. The overall privacy budget $(\varepsilon,\delta)$ remains exactly that of DPSGD and \textsc{DiSK}, guaranteed by the post-processing property of differential privacy, which ensures that any transformation applied after the privatization step cannot degrade the original privacy guarantees.

\section{Theoretical Analysis: Privacy-Utility Trade-off}

\label{ssec:privacy}

We theoretically analyze our method based on KF-filter method for differential privacy~\cite{ma2022kalman}. 

Let the FFT operator be $\mathcal F:\mathbb{R}^d\!\to\!\mathbb{C}^d$ with inverse
$\mathcal F^{-1}$, as introduced in Section~\ref{ssec:fft_recap}.
Fix a pivot index $k_0=\lfloor\lambda d\rfloor$ ($\lambda\!\in\!(0,1)$)
and a high–frequency attenuation ratio $\rho\in(0,1)$.
Define the diagonal spectral mask:
\begin{align}
  \Phi_\rho = \operatorname{diag}\!\bigl(
       \underbrace{1,\dots,1}_{k_0},
       \underbrace{1-\rho,\dots,1-\rho}_{d-k_0}
  \bigr),
\end{align}
and the deterministic post‑processing map $P(g)=\mathcal F^{-1}\!\bigl(\Phi_\rho\,\mathcal F g\bigr).$
Given a privatised gradient $g_t$, the filtered release is $\hat g_t := P(g_t).$ Then, privacy is preserved. \\

\noindent \textbf{Proposition 1: Post‑processing invariance} 
\textit{Because $P$ is data independent, $\hat g_t$ is
$(\varepsilon,\delta)$‑DP whenever the DiSK gradient
$g_t$ is $(\varepsilon,\delta)$‑DP.%
\footnote{This follows from the post‑processing theorem of differential
privacy~\cite[Thm.~2.1]{dwork2014algorithmic}.}} \label{prop:postprocess}
\\

The mask satisfies $\|\Phi_\rho\|_2=1$; hence
$P$ does not increase the $\ell_2$‑sensitivity of its input.
The Gaussian noise scale~$\sigma_w$ chosen for DiSK therefore
continues to satisfy the target $(\varepsilon,\delta)$ budget.
Consequently, Algorithm~\ref{alg:fftkf} inherits exactly the same
overall $(\varepsilon,\delta)$ guarantee as standard DP‑SGD/DiSK, computed
with the moments accountant over $T$ iterations.

\noindent \textbf{Lemma 1: Effect of the low‑pass mask}
\label{lem:covbias}
Write $g_t=\nabla F(x_t)+\eta_t$ with
$\eta_t\sim\mathcal N(0,\sigma_w^{2}I_d)$.
Let $\rho^\star=\bigl(k_0+(1-\rho)^{2}(d-k_0)\bigr)/d$.
Then
\[
  \mathbb E[\hat g_t]=A\,\nabla F(x_t),\qquad
  \mathrm{Cov}[\hat g_t]=\sigma_w^{2}A^{2},
\]
where $A:=\mathcal F^{-1}\Phi_\rho\,\mathcal F$ satisfies
$\|A-I_d\|_2 = \rho$ and
$\operatorname{tr}\!\bigl(\mathrm{Cov}[\hat g_t]\bigr)=
  \rho^\star\,d\sigma_w^{2}$. \\

\noindent\textit{Proof.} Unitary invariance of $\mathcal F$ yields the stated mean and covariance. Eigenvalues of $A$ are $1$ (multiplicity $k_0$) and $1-\rho$
(multiplicity $d-k_0$).

\noindent \textit{Remark.}\;
“Bias’’ in Lemma~\ref{lem:covbias} refers to $E[\hat g_t]-\nabla F(x_t)$;
filtering does not introduce systematic noise bias but scales the signal
by $A$.

Lemma~\ref{lem:covbias} replaces the isotropic noise term
$d\sigma_w^{2}$ in the DiSK analysis with $\rho^\star d\sigma_w^{2}$
and introduces a multiplicative bias factor $1-\rho$.
Repeating the steps of Theorem 2~\cite{zhang2025disk} yields:

\noindent \textbf{Theorem 2. Privacy–utility with FFT filtering}
\label{thm:pu-fft}
Under Assumptions \textsc{A1–A3} and the same
$(\eta,\kappa,\gamma)$ schedule as in Algorithm~\ref{alg:fftkf} satisfies
\begin{align}
  \frac1T &\sum_{t=0}^{T-1}\mathbb E\|\nabla F(x_t)\|^{2} \\
  &\le
  \frac{2\bigl(F(x_0)-F^\star + \beta\|\nabla F(x_0)\|^{2}\bigr)}
       {C_1\eta T}\\[4pt]
  &\qquad+\frac{2(\beta+\eta^{2}L)\kappa^{2}}{C_1\eta}
       \!\Bigl[(2+|1+\gamma|)\rho^\star\,d\sigma_w^{2}
              +\tfrac{\sigma_{\!SGD}^{2}}{B}\Bigr]\\[4pt]
  &\qquad+\rho^{2}\,G_T,
\end{align}
where
$G_T=\tfrac1T\sum_{t}\mathbb E\|\nabla F(x_t)\|^{2}$ and
\begin{align}
  C_1 = (1+\kappa-2\eta L)
        -4(\beta+\eta^{2}L)(1-\kappa)^{2}L^{2}\eta(2+|1+\gamma|).
\end{align}

\noindent\textbf{Practical choice and independence of the mask.}
In all experiments we fix $\lambda=\tfrac12$ and $\rho=0.5$
\emph{a priori} (i.e.\ independently of any individual training sample);
this gives $\rho^\star=0.625$ and $\rho^{2}=0.25$.
Thus the DP‑noise contribution is reduced by $37.5\%$
while the extra bias inflates the optimization term by at most $25\%$,
yielding a provably tighter trade‑off than plain DiSK.

\section{Experimental Results}

\begin{figure*}[t]
\centering
\begin{minipage}{0.3\textwidth}
\centering
\tiny
\begin{tabular}{llcccc}
\toprule
\textbf{Model} & \textbf{Method} & \textbf{CIFAR10} & \textbf{CIFAR100} & \textbf{MNIST} & \textbf{Tiny-ImgNet} \\
\midrule
\multirow{3}{*}{CNN5}  
  & DPAdam   & 60.83 & 23.34 & 91.18 & 10.93 \\
  & DISK     & 70.57 & 39.62 & 92.52 & \textbf{23.45} \\
  & FFTKF    & \textbf{70.86} & \textbf{40.80} & \textbf{92.62} & 22.62 \\
\midrule
\multirow{3}{*}{WRN-16}
  & DPAdam   & 53.81 & 15.53 & 91.18 & 10.63 \\
  & DISK     & 69.33 & 35.19 & 92.52 & 24.13 \\
  & FFTKF    & \textbf{70.38} & \textbf{35.91} & \textbf{92.62} & \textbf{24.61} \\
\midrule
\multirow{3}{*}{WRN-28}
  & DPAdam   & 53.72 & 14.74 & 91.18 & 9.94 \\
  & DISK     & 71.22 & 37.79 & 92.52 & 26.96 \\
  & FFTKF    & \textbf{72.58} & \textbf{38.93} & \textbf{92.62} & \textbf{27.63} \\
\midrule
\multirow{3}{*}{WRN-40}
  & DPAdam   & 54.50 & 14.05 & 91.18 & 8.92 \\
  & DISK     & 72.13 & 37.31 & 92.52 & 27.41 \\
  & FFTKF    & \textbf{73.73} & \textbf{37.95} & \textbf{92.62} & \textbf{27.83} \\
\midrule
\multirow{3}{*}{ViT-small}
  & DPAdam   & 50.98 & 19.83 & 91.18 & 13.47 \\
  & DISK     & 58.79 & 32.44 & 92.52 & 25.70 \\
  & FFTKF    & \textbf{59.85} & \textbf{32.46} & \textbf{92.62} & \textbf{25.96} \\
\bottomrule
\end{tabular}
\end{minipage}%
\hfill
\begin{minipage}{0.44\textwidth}
\centering
\includegraphics[width=\textwidth]{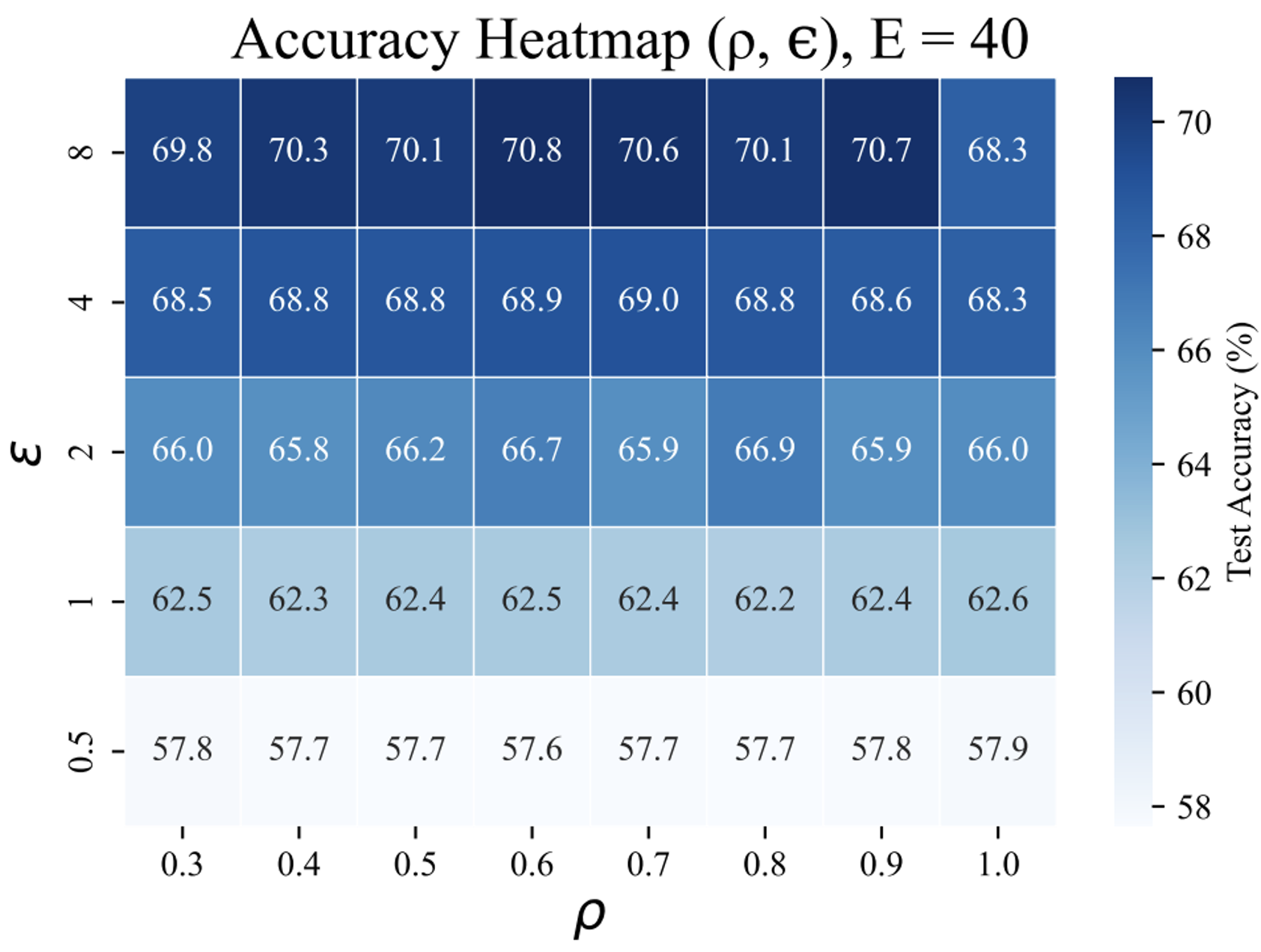}
\end{minipage}

\caption{%
Left: Test accuracy (\%) under $(\epsilon = 4)$ across four datasets and five model architectures. 
Right: Test accuracy across $(\rho, \epsilon)$ at epoch 40.}
\label{tab:fftkf_results_compact}
\end{figure*}

\label{sec:experiments}
In this section, we explore how the FFT-Enhanced Kalman Filter (FFTKF) improves the performance of differential privacy (DP) optimizers on various models, datasets, and privacy budgets. The utilization of FFT for the purpose of reshaping the DP noise in the frequency domain is undertaken with the objective of preserving the essential low-frequency gradient signal, while concomitantly directing privacy noise into spectral regions.

\medskip
\begin{figure*}[t]
    \centering
    \includegraphics[width=1\textwidth]{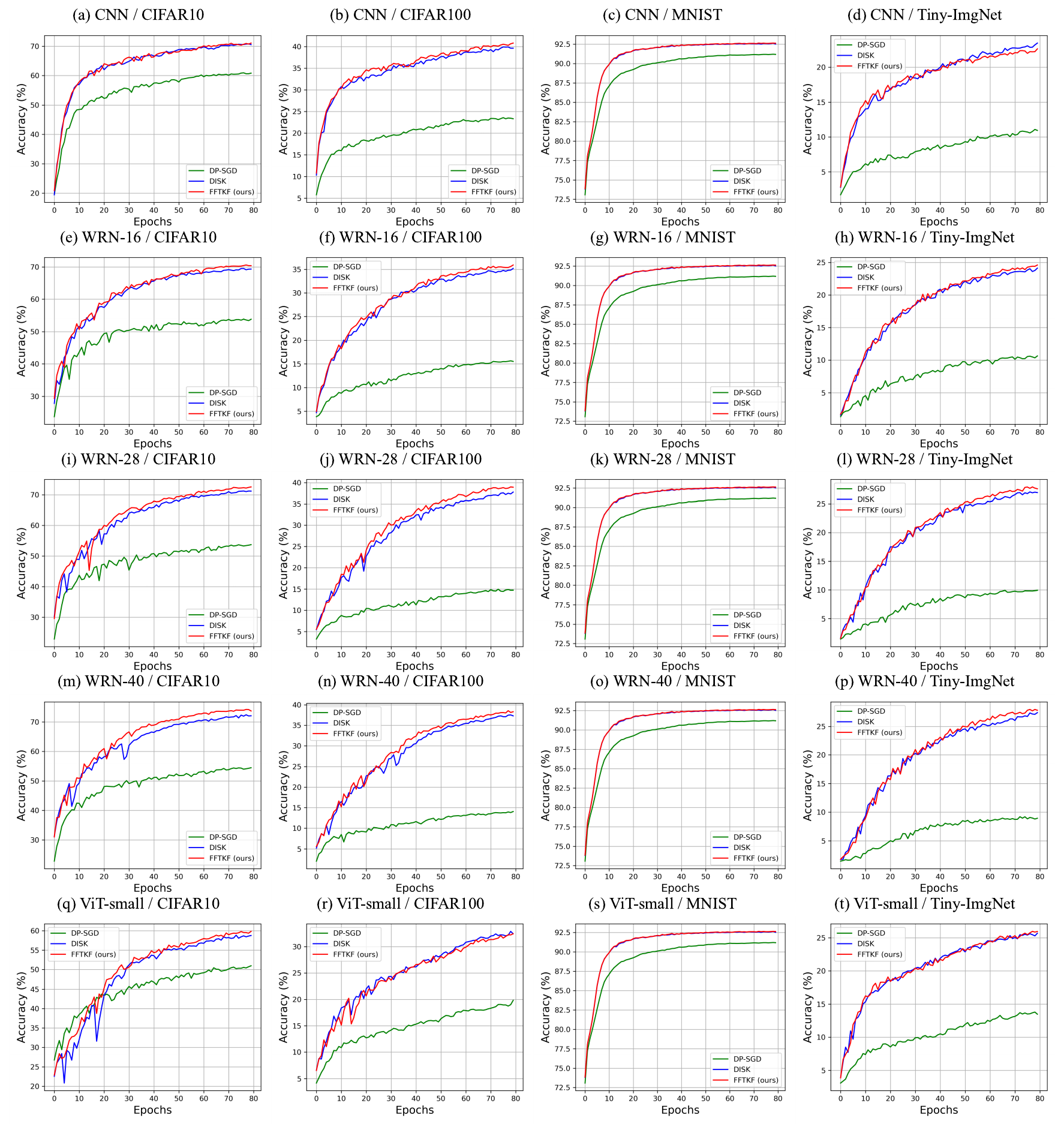}
    \vspace{-18pt}
    \caption{%
    Test accuracy curves at $\epsilon = 4$ across four datasets (CIFAR10~\cite{krizhevsky2009learning}, CIFAR100~\cite{krizhevsky2009learning}, MNIST~\cite{lecun1998gradient}, Tiny-ImageNet~\cite{le2015tiny}) and five model architectures. 
    Each plot compares DPAdam~\cite{dpadam} (green), DISK~\cite{zhang2025disk} (blue), and the proposed FFTKF-DPAdam (red). 
    FFTKF consistently improves final test accuracy, particularly on CIFAR and Tiny-ImageNet benchmarks.%
    }
    \label{fig:all_model_results}
    \vspace{-13pt}
\end{figure*}

\subsection{Experimental Settings}
The experiments are conducted on four standard image classification benchmarks, including MNIST\cite{lecun1998gradient}, CIFAR-10, CIFAR-100\cite{krizhevsky2009learning} and Tiny-ImageNet\cite{le2015tiny}. The experiments are conducted on three image classification models, including 5-layer CNN~\cite{krizhevsky2017imagenet}, Wide ResNet ~\cite{zagoruyko2016wide}, and ViT~\cite{dosovitskiy2020image}. A comparative analysis was conducted to assess the impact of FFTKF on various base algorithms, including the DP versions of Adam and SGD. The updates of these algorithms are delineated in Algorithm 1. In our experiments, the term \textit{FFTKF-} is employed to denote the privatized version of the FFT-enhanced Kalman filter algorithms.We apply a high-frequency shaping mask with parameters \( \rho \), where \( \rho \in (0, 1) \), to push DP noise into high-frequency components while preserving the essential low-frequency gradient signal. The pivot index \( k_0 \) is determined by the parameter \( \lambda \in (0, 1) \), which defines the transition point between low and high frequencies. In addition, we experimentally adjust the batch size \( B \), the total epochs \( E = \frac{N T}{B} \), and the learning rate \( \eta \) to achieve optimal performance within a given privacy budget \( \varepsilon \). The privacy parameter \( \delta \) is constant throughout all experiments to ensure a reasonable privacy guarantee.

\subsection{Numerical Results}
When operating within identical privacy budgets, the FFTKF consistently exhibits superior performance compared to baseline DP optimizers, including DPAdam and DISK, across a wide range of datasets and models. For example, when applied to CIFAR-10 with Wide ResNet-40, FFTKF demonstrates a test accuracy enhancement of up to 1.6\% over the best-performing state-of-the-art algorithm. On Tiny-ImageNet with ViT-small, FFTKF exhibits superior convergence stability and accuracy, a benefit that can be attributed to its effective spectral noise shaping.

\noindent As illustrated in Figure~\ref{fig:all_model_results} and Table~\ref{tab:fftkf_results_compact}, FFTKF achieves a better final precision within fixed privacy budgets. The efficacy of these enhancements is particularly evident under tight privacy constraints, where conventional optimizers frequently encounter significant noise corruption. The findings indicate the effectiveness of frequency domain filtering and Kalman-based prediction in mitigating the adverse effects of DP noise, particularly in high-dimensional vision tasks. \\

\begin{figure*}[t]
    \centering
    \includegraphics[width=1\textwidth]{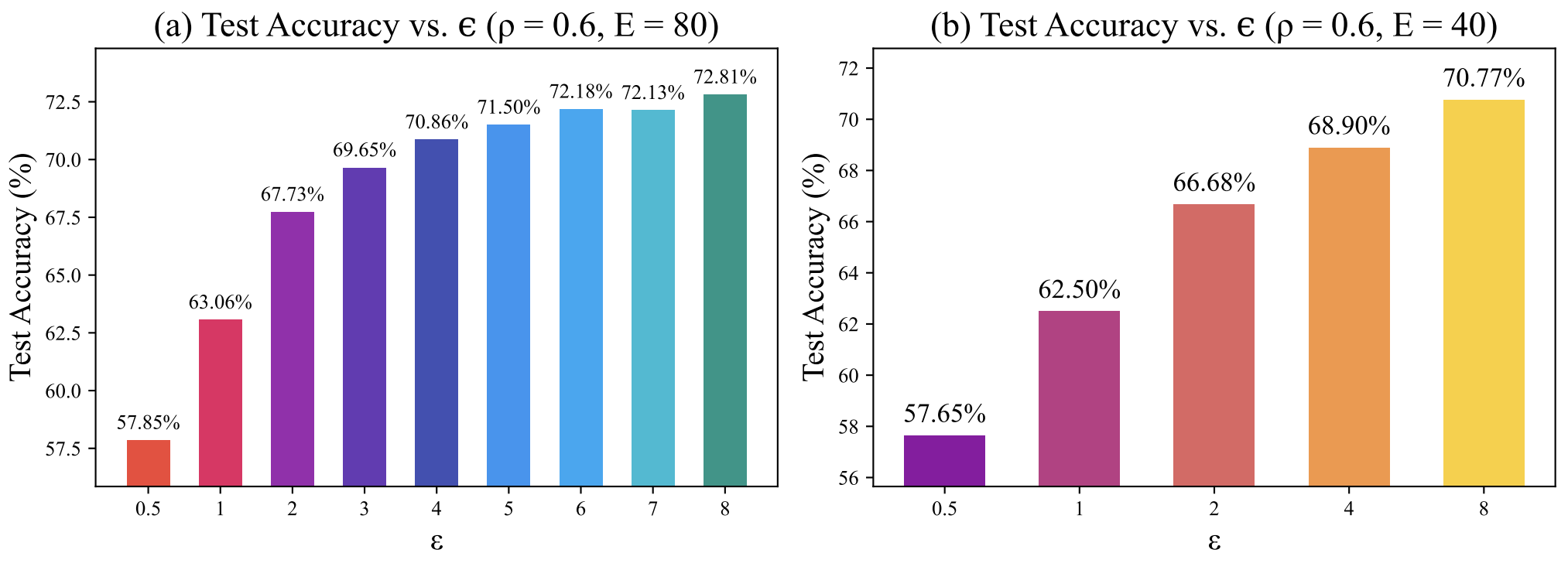}
    \vspace{-18pt}
    \caption{%
    Ablation study of FFTKF. 
            \textbf{(a)} Varying $\epsilon$ at epoch 80. 
            \textbf{(b)} Varying $\epsilon$ at epoch 40. 
    }
    \label{fig:fftkf_ablation}
\end{figure*}

\noindent\textbf{Ablation study.} To better understand the influence of FFTKF parameters, we conduct ablation studies on the high-frequency shaping parameter \( \rho \) and the privacy budget \( \epsilon \). We observe that moderate values of \( \rho \in [0.6, 0.7] \) provide a good trade-off between stability and adaptability. Furthermore, Figure~\ref{fig:fftkf_ablation} shows the result that higher values of \( \epsilon \), which imply weaker privacy but less noise, result in more accurate gradient estimation. The parameter \( \rho \) controls the redistribution of spectral noise and setting \( \rho = 0.6 \) consistently yields strong performance across a wide range of datasets.

\medskip

\section{Conclusion}
This paper introduced the FFT-Enhanced Kalman Filter (FFTKF), a differentially private optimization method that integrates frequency-domain noise shaping with Kalman filtering to enhance gradient quality while preserving $(\varepsilon, \delta)$-DP guarantees. By using FFT to concentrate privacy noise in high-frequency spectral components, FFTKF retains critical low-frequency gradient signals, complemented by a scalar-gain Kalman filter for further denoising. With a per-iteration complexity of $\mathcal{O}(d \log d)$, FFTKF demonstrates superior test accuracy over DP-SGD and DiSK across standard benchmarks, particularly under tight privacy constraints. Theoretically, FFTKF maintains equivalent privacy guarantees while achieving a tighter privacy-utility trade-off through reduced noise and controlled bias. FFTKF represents a significant advancement in efficient and effective private optimization.

\section{Acknowledgement}
This research was supported by Brian Impact Foundation, a non-profit organization dedicated to the advancement of science and technology for all.

\bibliographystyle{plain}
\bibliography{mybibfile}

\end{document}